\documentclass[letterpaper, 10 pt, conference]{ieeeconf}  
\usepackage{amsmath}
\usepackage{amssymb}
\usepackage{nccmath}
\usepackage{mathtools}
\newenvironment{mpmatrix}
{\begin{medsize}\begin{bmatrix}}%
    {\end{bmatrix}\end{medsize}}%
\usepackage[linesnumbered,ruled,vlined]{algorithm2e}
\usepackage{algorithmic}
\usepackage{graphicx}
\usepackage{graphics}
\usepackage{tikz}
\usetikzlibrary{calc,patterns,decorations.pathmorphing,decorations.markings}
\usetikzlibrary{positioning}
\makeatletter
\def\BState{\State\hskip-\ALG@thistlm}
\makeatother 
\graphicspath{{images/}}
\usepackage{textcomp}
\usepackage{booktabs}
\newcommand\numberthis{\addtocounter{equation}{1}\tag{\theequation}}

\IEEEoverridecommandlockouts                              
\title{\LARGE \bf
Exploiting the Natural Dynamics of Series Elastic Robots \\ by Actuator-Centered Sequential Linear Programming
}

\author{Rachel Schlossman$^{1}$, Gray C. Thomas, Orion Campbell, and Luis Sentis
\thanks{*This work was supported by NASA Space Technology Research Fellowships NNX15AQ33H (G.C.T) and 80NSSC17K0188 (R.S.), Office of Naval Research, ONR Grant N000141512507, and SOCOM STTR H92222-17-C-0050.}
\thanks{$^{1}${\tt\small rachel.schlossman@utexas.edu}}%
\thanks{Authors are with The Departments of Mechanical Engineering (R.S., G.C.T., O.C.) or Aerospace Engineering (L.S.), University of Texas at Austin, Austin, TX 78712-0292, USA}%
}

\begin{document}

\maketitle
\thispagestyle{empty}
\pagestyle{empty}


\begin{abstract}
Series elastic robots are best able to follow trajectories which obey the limitations of their actuators, since they cannot instantly change their joint forces. In fact, the performance of series elastic actuators can surpass that of ideal force source actuators by storing and releasing energy. In this paper, we formulate the trajectory optimization problem for series elastic robots in a novel way based on sequential linear programming. Our framework is unique in the separation of the actuator dynamics from the rest of the dynamics, and in the use of a tunable pseudo-mass parameter that improves the discretization accuracy of our approach. The actuator dynamics are truly linear, which allows them to be excluded from trust-region mechanics. This causes our algorithm to have similar run times with and without the actuator dynamics. We demonstrate our optimization algorithm by tuning high performance behaviors for a single-leg robot in simulation and on hardware for a single degree-of-freedom actuator testbed. The results show that compliance allows for faster motions and takes a similar amount of computation time.

\end{abstract}
\section{INTRODUCTION}

Since its inception \cite{pratt1995series}, a primary drawback of series elastic actuation has been the additional challenge for the control system. Human-centered robots commonly make use of series elastic actuators (SEAs), which offer the benefits of compliance---for safe interaction with humans---increased robustness, and force sensing \cite{paine2014design}. The compliant element is able to store and release energy, like human muscles, presenting an opportunity for increased efficiency and agility as compared to rigid actuators \cite{sreenath2011compliant}. Both feedback controllers and trajectory planners are faced with a more complex challenge when interfacing with these systems, yet modern control systems for human-centered robots (e.g., \cite{pratt2012capturability}) rely on a force-control planning abstraction which specifies an unmeetable goal for the low level feedback controller and provides those controllers with planned trajectories that do not respect their dynamic limitations. Our work addresses some of these issues.




Interest in modified series elastic actuators with clutches and variable stiffness compliant elements has driven many groups to derive bang-bang style and cyclic optimal behaviors to illustrate improved mechanical performance \cite{vanderborght2013variable}. Few groups, however, have investigated more general behaviors that allow for nonlinearities in the system. In \cite{chen2013optimal}, a convex optimization problem is formulated to maximize joint velocity by computing the switching times between rigid and compliant actuator behavior via the use of a clutch, but the actuator dynamics are linear except at switching times. One of the contributions of our work is the ability to handle the nonlinearities that are introduced at all time steps through a nonlinear transmission, while still leveraging compliance.

Iterative regulator-based optimal control has been successful in handling nonlinearities in these systems and achieving rapid motions in compliant robots, but is restricted in capturing state and input constraints, e.g., transmission speed or spring deflection limits. The iLQR indirect method has been modified to allow input constraints \cite{radulescu2012exploiting}, \cite{braun2012optimal}---but not state constraints directly. In \cite{braun2012optimal}, iLQR is used in combination with variable stiffness actuators to leverage the energy storing capability of the compliant element to throw a ball, but the motor position constraint can only be captured indirectly through the input constraint. Inequality state constraints in \cite{braun2013robots} are reformulated as canonical input constraints, yet the number of constraints possible with this strategy is at most the number of inputs. In contrast, our work captures all linear state and input constraints, which are upheld by the linear program.   

Spline-parameterized, nonlinear programming (NLP) and collocation approaches, based on general purpose large-scale NLP libraries, have been successfully applied to series elastic robots.
In \cite{werner2017optimal}, optimal walking trajectories are produced via NLP to be consistent with compliant dynamics subject to all relevant constraints with pre-defined contact transition times. 
\cite{werner2017generation} adds a collocation method to automatically select contacts, to automatically generate multiple steps of walking, and to jump, at the cost of approximating some actuator constraints. 
This approach leverages powerful and highly general NLP libraries, however, these general solvers result in long run-times on the order of an hour, even for problems that have roughly the same number of trajectory parameters as ours\footnote{1,782 parameters in ``less than an hour'' \cite{werner2017generation} versus our 1,176 parameters in 28.5 seconds for a two-link leg---iterating an LP 19 times.}.

In this paper, we propose a direct optimization algorithm which efficiently considers the nonlinear effects of the transmission linkages, robot dynamics, input and state constraints, and the energy storing capabilities of the series elastic elements. The algorithm uses sequential linear optimization to minimize a final velocity objective (with a 1-norm input penalty) to demonstrate its ability to produce high performance behaviors while satisfying system constraints. We formulate the problem as input selection for a time-varying discrete time linear system approximation that is updated iteratively. We formulate the system dynamics to connect the actuator space to the joint space. One of the key, novel features of our approach is the use of a fictitious pseudo-mass to improve discretization accuracy for the actuator component at large time steps. We find the pseudo-mass' value must be close to the reflected robot inertia to minimize simulation error and eigenvalue approximation error. A pseudo-mass of 0 kg results in unacceptable discretization inaccuracy. By exploiting problem structure via separating the linear and nonlinear components of our model, we typically achieve convergence within 20 iterations for a two-link system. Convergence is achieved more quickly when we test our approach on hardware for a single degree-of-freedom testbed. Our experiments demonstrate a greater degree of dynamic consistency and the leveraging of compliance when the spring dynamics are considered for trajectory generation.

\section{Modeling}\label{sec:modeling}

\subsection{Actuator Dynamics}\label{actuatordynamics}
Our model considers internal actuator dynamics, which are common for control design, but rare for trajectory design due to computational complexity. We follow the advice of \cite{orekhov2015unlumped} and \cite{schutz2016intuitive}, and connect three second-order systems through a differential to develop an unlumped model of the SEA. 

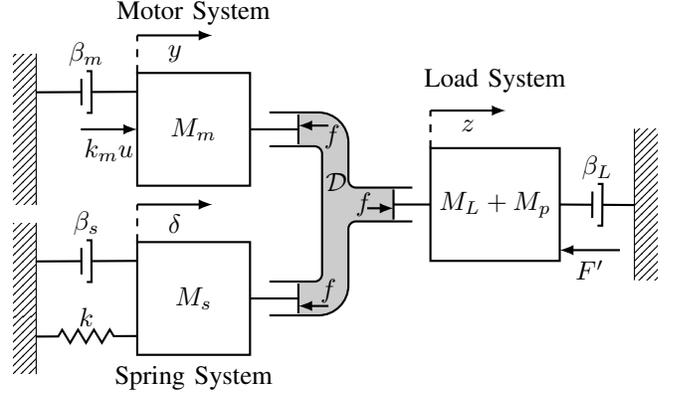
\begin{figure}
\begin{tikzpicture}

 \tikzstyle{spring}=[thick,decorate,decoration={zigzag,pre length=0.3cm,post
 length=0.3cm,segment length=6}]

 \tikzstyle{damper}=[thick,decoration={markings,  
   mark connection node=dmp,
   mark=at position 0.5 with 
   {
     \node (dmp) [thick,inner sep=0pt,transform shape,rotate=90,minimum
 width=15pt,minimum height=3pt,draw=none] {};
     \draw [thick] ($(dmp.north east)+(2pt,0)$) -- (dmp.south east) -- (dmp.south
 west) -- ($(dmp.north west)+(2pt,0)$);
     \draw [thick] ($(dmp.north)+(0,-5pt)$) -- ($(dmp.north)+(0,5pt)$);
   }
 }, decorate]

 \tikzstyle{ground}=[fill,pattern=north east lines,draw=none,minimum
 width=0.75cm,minimum height=0.3cm]

 \node[draw,outer sep=0pt,thick] (Mm) at (3,0) [minimum width=1.5cm, minimum height=1.5cm] {$M_m$};
 
 \node[yshift = .8cm]  at (Mm.north){Motor System};
 
  \node[draw,outer sep=0pt,thick] (Ms) [minimum width=1.5cm, minimum height=1.5cm, yshift = -1.5cm] at (Mm.south){$M_s$};
 
\node[yshift = -.3cm]  at (Ms.south){Spring System};
 
\node[outer sep=0pt,thick] (diff) [minimum width=.5cm, minimum height=4cm, xshift = 1.4cm, yshift = -1cm] at (Mm.east){$\mathcal D$};  

\fill[fill=gray!40] ($(diff.east) + (.25,.225)$) 
-- ($(diff.east) + (-.25,.225)$) 
arc (270:180:.1) 
-- +(0,.50)
arc (0:90:.4) 
-- +(-.25, 0)
-- +(-.25, -.45) 
-- ($(diff.east)+(-1.0,.775)$)
-- ($(diff.east)+(-.8,.775)$)
arc (90:0:.1)
-- +(0,-1.6)
arc (0:-90:.1)
-- ($(diff.east)+(-1.0,-1.025)$)
-- +(0,-.45)
-- +(.25,-.45)
arc (270:360:.4)
-- +(0,.75)
arc (180:90:.1)
-- +(.5,0)
;
\node[outer sep=0pt,thick] (difflabel) [minimum width=.5cm, minimum height=4cm, xshift = 1.15cm, yshift = -.7cm] at (Mm.east){$\mathcal D$};  

\node[draw,outer sep=0pt,thick] (Ml) [minimum width=1.7cm, minimum height=1.5cm, xshift = 1.6cm] at (diff.east){$M_L+M_{p}$};

 \node[yshift = .9cm]  at (Ml.north){Load System};

 \node (wall1) [ground,rotate=-90,minimum width=2cm, yshift = -1.5cm] at (Mm.west) {};
\draw (wall1.north east) -- (wall1.north west);
 
\node (wall2) [ground,rotate=-90,minimum width=2cm, yshift = -1.5cm] at (Ms.west) {};
\draw (wall2.north east) -- (wall2.north west);

\node (wall3) [ground,rotate=90,minimum width=2cm, yshift = -1.15cm] at (Ml.east) {};
\draw (wall3.north west) -- (wall3.north east);

 \draw[thick, -latex] ($(wall3.west) +  (-.35,.4)$) -- ($(wall3.west) +(-1.15,.4)$)
node [midway, below] {$F'$};  
 
 \draw[damper] ($(wall1.east) + (.15,1.5)$) -- ($(Mm.west) + (0,0.5)$)
 node [midway,yshift = .5cm] {$\beta_m$}; 

 \draw[spring] ($(wall2.east) - (-.15,-0.5)$) -- ($(Ms.west) - (0,0.5)$) 
 node [midway,above] {$k$};
 \draw[damper] ($(wall2.east) + (.15,1.5)$) -- ($(Ms.west) + (0,0.5)$)
 node [midway,yshift = .5cm] {$\beta_s$}; 

 \draw[damper] ($(Ml.east) + (0,0)$) -- ($(wall3.west) + (-.15,1.0)$)
 node [midway,yshift = .5cm] {$\beta_L$}; 
 
 \draw [thick]($(Mm.east) + (0,0)$) -- ($(diff.west) + (-.5,1)$);
 \draw [thick]($(diff.west) + (-.5,1.2)$) -- ($(diff.west) + (-.5,.8)$);
 
\draw [thick]($(Ms.east) + (0,0)$) -- ($(diff.west) + (-.5,-1.25)$);
\draw [thick]($(diff.west) + (-.5,-1.45)$) -- ($(diff.west) + (-.5,-1.05)$);
  
\draw [thick]($(diff.east) + (.25,0)$) -- ($(Ml.west) + (0,0)$);
\draw [thick]($(diff.east) + (.25,.2)$) -- ($(diff.east) + (.25,-.2)$);

\draw[thick]($(diff.east) + (.5,.225)$) 
-- ($(diff.east) + (-.25,.225)$) 
arc (270:180:.1) 
-- +(0,.50)
arc (0:90:.4) 
-- +(-.65,0)
;
\draw[thick] ($(diff.east) + (.5,-.225)$) -- 
($(diff.east) + (-.25,-.225)$)
arc (90:180:.1)
-- +(0,-0.75)
arc (0:-90:.4)
-- +(-.65,0)
;

\draw[thick] ($(diff.east)+(-1.4,.775)$)
-- ($(diff.east)+(-.8,.775)$)
arc (90:0:.1)
-- +(0,-1.6)
arc (0:-90:.1)
-- +(-.6,0)
;

 \draw[thick, dashed] ($(Mm.north west)$) -- ($(Mm.north west) + (0,.5)$);
 \draw[thick, -latex] ($(Mm.north west) + (0,0.5)$) -- 
                            ($(Mm.north west) + (1,0.5)$)
                            node [midway, below] {$y$};
                            
  \draw[thick, dashed] ($(Ms.north west)$) -- ($(Ms.north west) + (0,.5)$);
 \draw[thick, -latex] ($(Ms.north west) + (0,0.5)$) -- 
                            ($(Ms.north west) + (1,0.5)$)
                            node [midway, below] {$\delta$};
                            
  \draw[thick, dashed] ($(Ml.north west)$) -- ($(Ml.north west) + (0,.5)$);
 \draw[thick, -latex] ($(Ml.north west) + (0,0.5)$) -- 
                            ($(Ml.north west) + (1,0.5)$)
                            node [midway, below] {$z$};
                            
\draw[thick, -latex] ($(diff.north west) + (-0.1,-0.95)$) -- 
                            ($(diff.north west) + (-0.5,-0.95)$)
                            node [xshift=.45cm, yshift=-.15cm] {$f$};
 \draw[thick, -latex] ($(diff.south west) + (-0.1,.65)$) -- 
                            ($(diff.south west) + (-0.5,.65)$)
                            node [xshift=.4cm, yshift=.2cm] {$f$};      

 \draw[thick, -latex] ($(diff.east) + (-0.1,-.05)$) -- 
                            ($(diff.east) + (.25,-.05)$)
                            node [yshift=.05cm, xshift=-.4cm] {$f$};   
                            
 \draw[thick, -latex] ($(Mm.west) + (-0.75,0)$) -- 
                            ($(Mm.west) + (0,0)$)                                          	                node [midway, below] {$k_mu$};   
 \end{tikzpicture}
   \vspace{-0.5cm}  
  \caption{Internal dynamics of the SEA for the three-mass, differential constraint model. While there are no fluids in the physical SEA system, a fluid differential is used as a metaphor for the real mechanical differential, to easily visualize that the back forces are equal and that the motions of the spring and motor subsystems are in series. A pseudo-mass term, $M_p$, is introduced to allow discretization with longer time steps.}
  \label{fig:actuatordynamics}
    \vspace{-0.5cm}  
\end{figure}

The actuator model, shown in Fig. \ref{fig:actuatordynamics}, comprises the spring system; the motor system with input current, $u$; and the load system. The states considered are spring displacement, $\delta$; spring velocity, $\dot{\delta}$; motor displacement, $y$; and motor velocity, $\dot{y}$. The variables $z$ and $\dot{z}$ correspond to total actuator length and velocity, respectively. The motor subsystem is reflected to prismatic motion through the transmission---hence, all parameters of the subsystems are in linear units. The three systems are connected through a three-way mechanical differential,  $\mathcal D$, which enforces the relationship:
\begin{equation}\label{rel1}
z = \delta + y. 
\end{equation}
 
\noindent The dynamics of the three subsystems are: 

\begin{equation}\label{equ:1}
M_s\ddot{\delta} + \beta_s\dot{\delta}+k\delta = -f,
\end{equation}

\begin{equation}\label{equ:2}
(M_L+M_{p})\ddot{z} + \beta_L\dot{z} = f-(F-M_p\ddot{z}),
\end{equation}

\begin{equation}\label{equ:3}
M_m\ddot{y} + \beta_m\dot{y} = k_mu - f.
\end{equation}

$M_s$, $M_L$, and $M_m$ are the masses of the spring, load, and motor systems, respectively; $\beta_s$, $\beta_L$, and $\beta_m$ are these systems' respective damping coefficients; $k$ is the spring constant; and $k_m$ is the reflected motor constant. The second input, $F$, is the force output from the actuator, which is used to link with the robot dynamics and the nonlinearities in the system. $M_{p}$ is a fictitious pseudo-mass, which will be used to tune the eigenvalues of the linear actuator system before discretization, as discussed in Section \ref{section:LP}.
We define $F'$ as:

\begin{equation}
F' \triangleq F-M_p\ddot{z}.
\end{equation}

The variable $f$ is equal to the back forces from the differential and, equivalently, the Lagrange multiplier which enforces the differential constraint. Substitution for $f$ reveals that this model is ultimately fourth order:  

\begin{equation}\label{eq:nonlin1}
E_o \dot{x} = A_o x + B_{o,u} u + B_{o,F} F', 
\end{equation}

\noindent  where state vector $x \triangleq \begin{bmatrix}
	\delta & \dot{\delta} & y & \dot{y}\\
\end{bmatrix}^{T}$ and

\begin{align*}
E_o \triangleq \begin{bmatrix}
    1 & 0 & 0 & 0 \\
    0 & M_s+M_L+M_{p} & 0 & M_L+M_p  \\
    0 & 0 & 1 & 0\\
    0 & M_L+M_p & 0 & M_m+M_L+M_{p} \\
\end{bmatrix}
\end{align*}

\begin{align*}
A_o \triangleq \begin{mpmatrix}
    0 & 1 & 0 & 0 \\
    -k & -(\beta_s + \beta_L) & 0 & -\beta_L \\
    0 & 0 & 0 & 1 \\
    0 & -\beta_L & 0 & -(\beta_L + \beta_m) \\
\end{mpmatrix}
\end{align*}


\begin{align*}
B_{o,u} \triangleq \begin{bmatrix}
	0 & 0 & 0 & k_{m}    
\end{bmatrix}^T, \quad
 B_{o,F} \triangleq \begin{bmatrix}
	0 & -1 & 0 & -1
\end{bmatrix}^T.
\end{align*}

\noindent Rearranging (\ref{eq:nonlin1}),

\begin{equation} \label{continuous}
\dot{x}
=
A_*x + B_{*,u}u + B_{*,F}F',
\end{equation}

\noindent where 

\begin{align*}
A_* \triangleq E_0^{-1}A_0, \quad B_{*,u} \triangleq E_0^{-1}B_{o,u}, \quad \text{and}
\end{align*}

\begin{align*}
B_{*,F} \triangleq E_0^{-1}B_{0,F}.
\end{align*}

From the construction of $A_0$, it is clear that the eigenvalues of $A_*$ will vary with $M_p$. As we proceed, we will discuss the application of this formulation for the general case of $p$ joints. Our state vector will be extended to:

\begin{equation}
x = [x_1^T, x_2^T,\dotsc, x_p^T]^T,
\end{equation}

\noindent where each $x_i$ captures the four states described in (\ref{eq:nonlin1}) for their respective actuator system. Equation (\ref{eq:nonlin1}) is extended (using the Kronecker product $\otimes$) to a $p$-link system with:  

\begin{equation}\label{eqn:general1}
E_{o,p} = I_p \otimes E_o, \quad A_{o,p} = I_p \otimes A_o, 
\end{equation}

\begin{equation}\label{eqn:general2}
B_{o,u,p} = I_p \otimes B_{o,u}, \quad \text{and} \quad B_{o,F,p} = I_p \otimes B_{o,F}, 
\end{equation}

\noindent where $I_p$ is the $p$x$p$ identity matrix. Equation (\ref{continuous}) can then be reformulated using (\ref{eqn:general1}) and (\ref{eqn:general2}) to obtain $A_1$, $B_{1,u}$, and $B_{1,F}$ for $p$ joints:

\begin{equation} \label{general_continuous}
\dot{x}
=
A_1x + B_{1,u}u + B_{1,F}F'.
\end{equation}

\subsection{Robot Dynamics}

The force $F$ connects the actuator to the robot dynamics. In general, for a multi-link system, the dynamics are:

\begin{equation}\label{equ:LagrangeDyn}
M(q)\ddot{q}+C(q,\dot{q})+G(q) = \tau = L(q)^TF,
\end{equation}

\noindent where $M$, $C,$ and $G$ represent inertia, Coriolis and centrifugal, and gravitational forces, respectively, and $q$ is the generalized joint angle vector.

The angle-dependent moment arm between the actuator and the joint, $L(q)$ abbreviated $L$, serves as the Jacobian between the joint space and the actuator space: $L\dot{q} = \dot z$. We solve for $F'$ by projecting it into the actuator-position--actuator-force space and manipulating (\ref{equ:LagrangeDyn}):

\begin{equation}
\label{eqn_F_prime_pre_zdd_subs}
F' = F-M_p\ddot{z} = (L^{-T}M(q)L^{-1}-M_p)\ddot{z}+b(q,\dot{q}),
\end{equation}
where

\begin{equation}
b(q,\dot{q}) \triangleq L^{-T}(C(q,\dot{q})+G(q)-M(q)L^{-1}\dot{L}\dot{q}).
\end{equation}

\noindent This is an expression for the impedance of the robot at the $\{ \dot{z},F'\}$ port. 










\subsection{Discretization}

To prepare for discrete time $u$ optimization, the state space model is discretized into N time steps of length $\Delta T$. By the continuous state space model in (\ref{general_continuous}), acceleration at the actuator output can be computed as:

\begin{equation}\label{approxZ}
\ddot{z} = 
S(A_1x +
B_1 \begin{bmatrix}
u\\
F'\\
\end{bmatrix}),
\end{equation}

\noindent where $B_1$ is the concatenation of $B_{1,u}$ and $B_{1,F}$. This is actuator admittance at the $\{ \dot{z},F'\}$ port. $S$ is formulated to capture the acceleration terms for the $p$-link system:

\begin{equation}
S =  I_p\otimes
\begin{bmatrix}
0 & 1 & 0 & 1
\end{bmatrix}.
\end{equation}

\noindent $F'$ is expressed in terms of the states by substituting (\ref{approxZ}) into (\ref{eqn_F_prime_pre_zdd_subs}):

\rule{0pt}{.05in}
\begin{equation}\label{eqn:F}
\begin{aligned}
F' = & [I-(L^{-T}M(q)L^{-1}-M_p)SB_{1,F}]^{-1}
[b(q,\dot{q})+\\
&+(L^{-T}M(q)L^{-1}-M_p)(SA_1x+SB_{1,u}u)].\\
\end{aligned}
\end{equation}

\noindent We discretize the linear actuator admittance model under the zero-order hold assumption for both $u$ and $F'$. The discrete state space model is then: 

\begin{equation}\label{equ:mainequation}
x_{n+1} = Ax_{n}+B\begin{bmatrix}
u_{n}\\
F'_{n}\\
\end{bmatrix}, 
\end{equation}
where  
\begin{equation}
A \triangleq e^{A_1\Delta T}, \quad B \triangleq \int_{0}^{\Delta T} e^{A_1(\Delta T-\tau)}B_1d\tau.
\end{equation}

We combine discrete time admittance and impedance at the $\{ \dot{z},F'\}$ interface by grouping terms which are linear in $x$ and $u$. The discretized (time-varying) update equation is: 



\begin{equation}\label{eq:discrete}
x_{n+1} = A_{lin,n}x_{n} + B_{lin,n}u_{n} + {bias}_{n}, 
\end{equation}
where $A_{lin}$ and $B_{lin}$ capture the linear dynamics associated with the actuator states and input current, respectively, and $bias$ captures the nonlinear robot impedance, including gravity, Coriolis effects, and nonlinear transmissions. The $M_p$ parameter is used to minimize the error introduced by discretizing the actuator admittance in the absence of the reflected inertia of the robot links. The $x$ and $u$ vectors at each time step are concatenated to form the trajectory matrices $\mathbf{X} \triangleq [x_1, x_2, ..., x_N]$ and $\mathbf{U} \triangleq [u_1, u_2, ..., u_{N-1}]$, respectively. This locally-linear model forms the foundation from which our algorithm is developed.  

\section{Iterative Linear Programming}\label{section:LP}

For trajectory optimization of a $p$-link system, our approach follows a strategy that culminates in a linear programming subproblem. Our local optimization approach requires a baseline trajectory, $\mathbf{Z}_{base}$, the concatenation of $z_{base}$ over all time steps, to initialize the nonlinear parts of the dynamics. A slow trajectory or a static position both serve as good choices. There is no need for a similar baseline trajectory for the actuator states due to our exploitation of their linear problem structure. The $\mathbf{Z}_{\mathrm{base}}$ trajectory allows us to compute the time-varying matrices used in \eqref{eqn:F} to compute $F'$. We can then compute the linearization components, $A_{lin}$, $B_{lin}$, and $bias$, for each time step (effectively saving our solver from eliminating the $F'$ variable itself). 

The linear problem structure can then be exploited. New displacement and velocity trajectories for the spring and motor subsystems are computed via a linear program and are captured in the optimal trajectory, $\mathbf{X}^*$. The optimal control parameters over all time steps, captured in $\mathbf{U}^*$, are also produced. The resulting $\mathbf{Z}$ trajectory becomes the new $\mathbf{Z}_{\mathrm{base}}$, and $\mathbf{X}^*$ is used to compute the new $F'$, $A_{lin}$, $B_{lin}$, and $bias$ matrices for the next iteration. Trust region constraints will keep the next $\mathbf{Z}$ trajectory close to this updated $\mathbf{Z}_{\mathrm{base}}$ trajectory. The algorithm continues to run until the 2-norm of the difference between the current and previous trajectories stops changing.

A key benefit of our approach is that all relevant actuator state and input constraints can be included in the formulation. The constraints are associated with the upper and lower bounds of the allowable spring deflections, $\overline{\delta}$, joint limits, actuator ballscrew velocity, $\overline{\dot{y}}$, and input currents, $\overline{u}$. The parameter $\overline{\Delta z}$ defines the trust region, which can be used to aid convergence of the iteration scheme. We note that the dimension of this trust region is small relative to the full dimension of $\mathbf{X}$---again due to separation of the linear and nonlinear dynamics. The final state can be subject to partial end point constraints. Our linear subproblem minimizes a problem-specific, linear cost function, $h(\mathbf{X},\mathbf{U})$, which is a function of, and is subject to linear constraints on, the discretized states and inputs:

\newcommand{\spacefix}{$\:$ }
\begin{align*}
& \underset{\mathbf{X}, \mathbf{U}, \mathbf{U}_{abs}}{\text{minimize }} 
    && h(\mathbf{X}, \mathbf{U}) \numberthis\label{eq:cvxsubprob} \\
  &  \text{subject to}&
& \text{dynamics:\ }
 \eqref{eq:discrete} &\forall\ & n \in \mathcal N/N \\
 \shortintertext{\spacefix a trust region:}
&&& |z_{i,n} - z_{i,n,base}| \leq \overline{\Delta z} &\forall\ &  i\in\mathcal P,\  n\in \mathcal N \\ 
\shortintertext{\spacefix state and input constraints:}
&&& |\delta_{i,n}|\leq \overline{\delta} &\forall\  & i\in\mathcal P,\  n\in \mathcal N \\
&&& z_{min,i} \leq z_{i,n} \leq z_{max,i} &\forall\  & i\in\mathcal P,\  n\in \mathcal N  \\
&&& |\dot{y}_{i,n}| \leq \overline{\dot{y}} &\forall\  & i\in\mathcal P,\  n\in \mathcal N \\
&&& |u_{i,n}| \leq \overline{u} &\forall\  & i\in\mathcal P,\  n\in \mathcal N/N   
\shortintertext{\spacefix and problem-specific constraints, in two our studies:}
&&& x_{1} = x_{\mathrm{init.}},\ z_N=z_{\mathrm{fin.}}\\
&&& |u_{dev,i,n}|\leq u_{abs,i, n} &\forall\  & i\in\mathcal P,\  n\in \mathcal N/N
\shortintertext{\spacefix and in our single-leg simulation only:}
&&& J_{\mathrm{com\_x\_velocity}}\dot z_N = 0\\
&&& \Phi_{i,n} \geq 0 &\forall\  & i\in 4,\  n\in \mathcal N/N \\
\end{align*}
where $|a|\leq b$ is shorthand for two linear inequalities, $-b\leq a\leq b$; $\forall \ n\in\mathcal N$ means $n=1,\dotsc,N$; $\forall\ i\in \mathcal P$ means $i=1,\dotsc,p$; and $/$ means omitting an element from the set. The parameter $\Phi$ refers to the foot contact constraints, which will be discussed in Section \ref{ground_contacts}. The variables $u_{dev}$ and $u_{abs}$ are used to minimize deviations from an equilibrium input trajectory, as discussed in Section \ref{section:velocitymaximization}. The specific cost functions used in our studies and problem-specific constraints are described further in the Simulation and Experiments sections.


To achieve convergence, we choose $\overline{\Delta z}$ to limit planning to the region where our linearized dynamics are not too inaccurate. Our novel approach is to select $M_p$ so that the fastest eigenvalue over the entire trajectory (which corresponds to the spring oscillation mode in the systems we studied) of $A_{lin}$ and $A_1$ approximates the fastest eigenvalue of the continuous system, ensuring an accurate approximation of the system dynamics. When $M_p = 0$, the spring dynamics settle faster than one time step and cannot be leveraged.


\section{Simulation}

\subsection{Apptronik\textsuperscript{TM} Draco-Inspired System}\label{robot_model}
\begin{figure}[tp]
\rule{0pt}{.1in}
\centering
\includegraphics[width=\columnwidth]{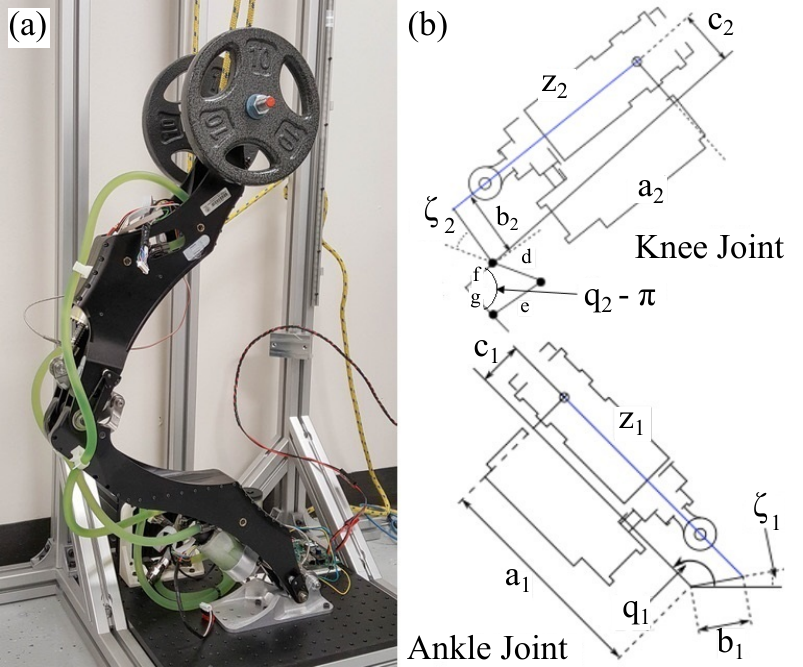}
\caption{(a) Draco leg prototype. Our simulation-only experiments are modeled after this robot, with significantly reduced spring rates, performing the liftoff phase of a jump. (b) Schematics emphasize the nonlinear transmissions between actuator length, $z$, and joint angle, $q$, for the ankle and knee. These nonlinear transmissions motivate our choice to represent the robot impedance in actuator-length--actuator-force space rather than the standard joint-angle--joint-torque space.}
\label{fig:legpic}
  \vspace{-0.7cm}  
\end{figure}

The formulation in the previous section is applied to the two-link Draco robot (Fig. \ref{fig:legpic}) in simulation. The Draco humanoid robot leg prototype is driven by viscoelastic actuators at its ankle and knee joints. Because the viscoelastic actuators used in the Draco system are very stiff, approximately $8e^{6}$ N/m, these elements offer minimal energy-storing capabilities. For this study, we explore the advantages of implementing softer springs in this system for a high-performance task.


 The state space model in (\ref{eq:discrete}) is used with $p = 2$. The two actuators have equivalent spring, motor, and load dynamics. The Draco leg, excluding the actuation linkages, is essentially a two-link manipulator. 
The process to develop the dynamic equations of the robot to include the actuator states follows that described in Section \ref{sec:modeling}. The variables $F'_1$ and $F'_2$ are obtained from Lagrangian dynamics with $M(q)\in R^{2\text{x}2}$, and $C(q, \ddot{q})$, $G(q)$ $\in R^{2}$. Due to space limitations, the coefficients of $M(q)$, $C(q,\dot{q})$, and $G(q)$ for this two-link robot can be found in \cite{al05}. 

To use (\ref{eqn:F}), the moment arms, $L_1(q_1)$ and $L_2(q_2)$, of the ankle and knee joints, respectively, must be considered as: 

\begin{equation} L \triangleq
\begin{bmatrix}
	L_1(q_1) & 0 \\
    0 & L_2(q_2) \\
\end{bmatrix}.
\end{equation}

We chose to demonstrate our algorithm for the goal of maximizing velocity at the center of mass (COM) of the robot, to obtain an optimal trajectory for a jumping motion. The parameters used for the simulation were guided by system identification of our lab's SEA and the parameters of the Draco leg. Select parameters are included in Table I. In the table, the parameters $I_{1}$ and $I_{2}$,  $m_{1}$ and $m_{2}$, and $l_1$ and $l_2$ equal the moments of inertia, masses, and lengths of the lower and upper legs, respectively. 
 \vspace{-0.2cm}  
\begin{table}[ht]
    \begin{minipage}{.36\columnwidth}
    \label{DracoParametersTable}
    \centering
      \caption{Dynamics}
        \begin{tabular}{r|l}
\toprule
$M_S$ (kg) & 1.7\\
$k_S$ (N/m) & 250k\\
$\beta_S$ (Ns/m) & 0\\
$M_m$ (kg) & 293\\
$\beta_m$ (Ns/m) & 1680\\
$M_L$ (kg) & 0\\
$M_p$ (kg) & 580\\
$\beta_L$ (Ns/m) & 0\\
$I_{1}$ (kg-$m^2$) & 0.077\\
$I_{2}$ (kg-$m^2$) & 0.050\\
$m_{1} (kg)$ & 3.77\\
$m_{2}$ (kg) & 15\\
$l_{1}$ (m) & 0.5\\
$l_{2}$ (m) & 0.5\\

        \end{tabular}
    \end{minipage}%
    \begin{minipage}{.3\columnwidth}
    \centering
    \caption{Transmissions (Fig. 2)}
        \begin{tabular}{r|l}
\toprule
$a_1$ (m) & 0.21\\
$b_1$ (m) & 0.04\\
$c_1$ (m) & 0.02\\
$\zeta_1$ (rad) & .464\\
$a_2$ (m) & 0.2\\
$b_2$ (m) & 0.05\\
$c_2$ (m) & 0.04\\
$d$ (m) & 0.04\\
$e$ (m) & 0.03\\
$f$ (m) & 0.03\\
$g$ (m) & 0.01\\
$\zeta_2$ (rad) & .524\\

        \end{tabular}
    \end{minipage}%
   \begin{minipage}{.33\columnwidth}
   \label{CostFunctionConstraints}
     \centering
    \caption{Constraints}
        \begin{tabular}{r|l}
\toprule
$\overline{\delta} $ (m) & 0.012\\
$z_{min,1}$ (m)& .1700 \\
$z_{min,2}$ (m) & .1563 \\
$z_{max,1}$ (m) & .2351\\
$z_{max,2}$ (m) & .2304\\
$\overline{\dot{y}}$ (m/s) & 0.3\\
$\overline{u}$ (A) & 15 \\
$\overline{\Delta {z}}$ (m) & 0.1 \\
$q_{1N}$ (rad) & 1.96\\
$q_{2N}$ (rad) & 5.30\\
$N$ & 85\\
$\Delta T$ (s) & .0095 \\

\end{tabular}
    \end{minipage}%
\end{table}

\subsection{Ground Contacts}\label{ground_contacts}
Ground contact wrenches are considered in the Draco model in the styles of \cite{koolen2016design}\footnote{In our 2D simulations, this style of linear parameterization is not an approximation of the true friction cone, but it is in 3D space.}, \cite{thomas2016towards}. Point contacts with static Coulomb friction, with the coefficient of friction, $\mu = 0.8$, are applied: one at the front of the foot and one at the heel. Friction cones are formulated at each contact point using the basis vectors $b_1 = \begin{bmatrix}
	\mu & 1 \\
\end{bmatrix}^{T}$ and $b_2 = \begin{bmatrix}
	-\mu & 1 \\
\end{bmatrix}^{T}$. The positive force intensity parameters $\Phi_1$, $\Phi_2$, $\Phi_3$, and $\Phi_4$ are the basis vector multipliers, with two of these force intensities associated with each end of the foot, as shown in Fig. \ref{fig:simulation}.a. Our linear program poses as equality constraints that the contact wrenches must satisfy Newton's second law in the x, y, and rotational directions. The force intensities must also be greater than or equal to zero until the robot jumps, as indicated in Section \ref{section:LP}. These constraints imply a zero moment point condition \cite{koolen2016design}.

\subsection{Velocity Maximization for Jumping}\label{section:velocitymaximization}

In this study, the cost function to be minimized expresses the goal to maximize the upward y-velocity of the robot COM at the final time, $V^* \triangleq J_{\mathrm{com\_y\_velocity}}\dot{z}_N$, where this Jacobian is known a-priori due to our constrained final position, $z_{\mathrm{fin.}}$.  The simulation mimics the configuration shown in Fig. \ref{fig:legpic}. We also strive to avoid unnecessary deviations from the motor current trajectory which keeps the robot at equilibrium with its springs, $\mathbf{U}_{baseline}$. We amend the cost function (to be minimized) to include the 1-norm of deviation from the baseline control signal, $\mathbf{U}_{dev} = \mathbf{U}-\mathbf{U}_{baseline}$. However, to keep the cost function linear, we create the variable matrix $\mathbf{U}_{abs}$ to represent $|\mathbf{U}_{dev}|$, as shown in (\ref{eq:cvxsubprob}). We have also added a slight preference towards solutions with small force intensities: 

\begin{equation}\label{costfunction}
\begin{aligned}
h(\mathbf{X}, \mathbf{U}) = &-J_{\mathrm{com\_y\_velocity}}\dot z_N
 +\alpha \sum_{i\in\mathcal P}\sum_{n\in\mathcal N/N} u_{abs,i,n} +\\ & + \gamma \sum_{i\in 4}\sum_{n\in\mathcal N/N} \Phi_{i,n}, 
\end{aligned}
\end{equation}

\noindent where $\alpha$ equals $1e^{-5}$ and $\gamma$ equals $1e^{-8}$. This cost function is linear, supporting our problem structure. Considering (\ref{eq:cvxsubprob}), $\Phi$ is also an optimization variable in this problem.

For our simulation, the initial condition is at equilibrium with the two springs, which drives the formulation of $\mathbf{Z}_{base}$. The initial and final conditions capture that the leg position starts and ends at the same angular configurations, $q_{1N}$ and $q_{2N}$. The final constraint is that the x-component of velocity at the COM is equal to zero at the final time. 

The sequential linear optimization problem is solved using the Matlab CVX library \cite{cvx} with the Gurobi solver. 
A time period of 0.798 s is considered. The algorithm converges in 19 iterations, $j = 19$, within a tolerance of 0.001 for $||\mathbf{X}^*_{j} - \mathbf{X}^*_{j-1}||_2$. The corresponding behavior is shown in Fig. \ref{fig:trajectories}.a-\ref{fig:trajectories}.c. An optimal value of 1.92 m/s upward velocity is achieved. One will notice spring oscillations, demonstrating the use of the two springs to store and release energy. Draco bends down and springs upward, following a jumping trajectory. Fig. \ref{fig:eigenvalues}.a demonstrates exponential convergence of our iteration scheme. 

\begin{figure}[tp]
\rule{0pt}{.1in}

\begin{center}
  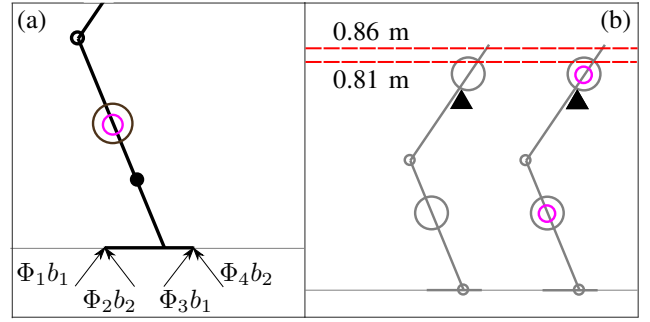
  \caption{The simulated robots. (a) Point contacts at the front (left) and back (right) of the foot. (b) The rigid robot, left, and compliant robot, right (with the springs indicated in pink), after they jump and return to the ground. The COMs of the two robots are illustrated as black triangles. The two COM initial heights are both 0.67 m when the robots lift into the air, and the maximum heights of the compliant and rigid configuration COMs are 0.86 m and 0.81 m, respectively, which are marked with red lines.}  \label{fig:simulation}
    \vspace{-0.5cm}  
\end{center}
\end{figure}

\begin{figure}[tp]

\def\svgwidth{.98\columnwidth}
  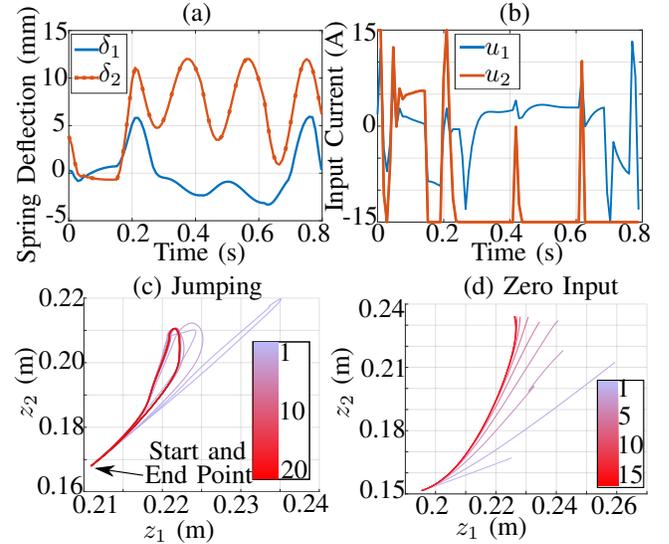
  \caption{(a) Spring deflection trajectories for the compliant leg's optimal behavior. (b) The corresponding optimal u's to produce the optimal trajectory, which operate at the input limits. (c) The z trajectories produced over 20 iterations to produce the jumping behavior, demonstrating convergence. (d) The z trajectories produced over 15 iterations ($M_p$ = 580 kg and $\Delta$T = .0095 s), showing convergence for the system's zero input behavior.}
  \label{fig:trajectories}
    \vspace{-0.75cm}  
\end{figure}

\begin{figure}
\rule{0pt}{.1in}

  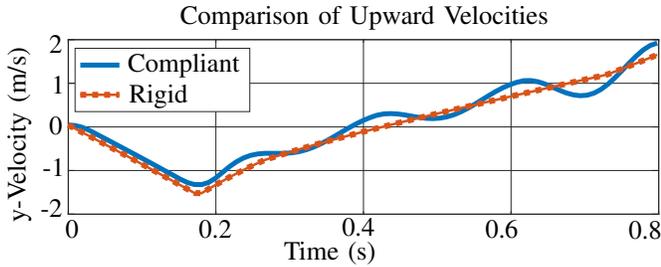
  \caption{This comparison between the upward velocity components of the optimal trajectories for the rigid and compliant systems shows that the final velocity of the compliant system is 1.2 times that of the rigid system.}  \label{fig:upward_velocity_comparison}
    \vspace{-0.25cm}
\end{figure} 

This problem can also be formulated with the assumption of rigid actuators to allow for a direct comparison between the optimized trajectories for the rigid and compliant cases. Specifically, (\ref{eq:nonlin1}), (\ref{eqn:general1}), and (\ref{eqn:general2}) are used without considering the spring subsystem. The cost function in (\ref{costfunction}) is used for the rigid and compliant cases, and the resulting optimal motions are compared. Considering the same initial heights of the robots' COMs, the compliant leg's COM reaches a height that is 36\% higher than that of its rigid counterpart. Fig. \ref{fig:simulation}.b shows the  associated Matlab simulation with a comparison of the achieved COM heights. Fig. \ref{fig:upward_velocity_comparison} shows that the optimal velocity in the compliant configuration, 1.92 m/s, is 16\% greater than that of the the rigid configuration, 1.65 m/s. For the rigid robot, the ball screw limits are not reached, but its motion is still constrained by acceleration limits, damping, and ground contact constraints.  These results demonstrate the gains that can be achieved from leveraging the dynamics of the springs. 

Our optimization program for the rigid system converges in 25 iterations, as compared to 19 iterations in the compliant simulation. Table IV shows the breakdown of average computation time per iteration to calculate $F'$, $A_{lin}$, $B_{lin}$, $bias$, and the wrench components, and the time spent in the Gurobi optimizer. These results demonstrate that consideration of compliance introduces only slightly increased computational costs in our method.  

\begin{table}[ht]
\caption{Average Time per Iteration for Algorithm Components and Total Simulation Time (s) } 
\centering 
\begin{tabular}{c c c c} 
\hline\hline 
Configuration & Linearization & Optimization & Total Time \\ [0.5ex] 
\hline 
Compliant & 0.077 & 1.32 & 28.5 \\ 
Rigid & 0.072 &  1.14 & 32.1 \\ 
\hline 
\end{tabular}
\label{table:nonlin} 
      \vspace{-0.15cm}
\end{table}


\begin{figure}[tp]

\def\svgwidth{.98\columnwidth}
  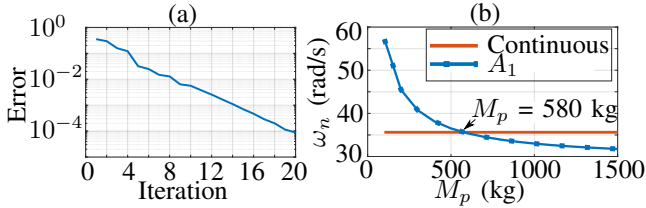
  \caption{(a) With the 25th iteration trajectory from the compliant jumping study as a baseline, the results suggest that the error decays exponentially. (b) The natural frequency of $A_1$ (linear time-invariant actuator dynamics with pseudo-mass) varies with $M_p$, and we select an $M_p$ value where the frequency aligns with that of the nonlinear, continuous dynamics at 35 rad/s.}
  \vspace{-0.5cm}
  \label{fig:eigenvalues}
\end{figure}



\subsection{Zero Input Behavior}
To validate simulation accuracy, we ensured that energy was conserved throughout a zero input simulation. A test was conducted in which the system was released from rest from a nearly vertical position. The motors were off and no current was sent to the system. With $\Delta T = 0.0095$ s, energy varies by 1.79\% with the reasonable pseudo-mass, 580 kg. 

Compared to the jumping studies, the system was more heavily influenced by the changing transmission as the links fell downward due to gravity. As shown in Fig. \ref{fig:trajectories}.d, the algorithm converged quickly, in 12 iterations, even in this highly nonlinear case, thereby demonstrating its success in handling nonlinearities in the system.


\subsection{Pseudo-Mass Selection} \label{pseudo_selection}

The spring oscillation eigenvalue of the actuator system is influenced by the reflected link inertia, and can exceed the sampling rate (and therefore suffer from aliasing when discretized) in the absence of a tuned pseudo-mass parameter. $M_p$ is set to 580 kg for the two actuators based on closeness to the largest eigenvalue in the expected operational range. 

Fig. \ref{fig:eigenvalues}.b illustrates the importance of selecting a reasonably-tuned $M_p$ value. The `Continuous' eigenvalues, which are independent of $M_p$, represent the full dynamics of the nonlinear system. With a tuned value of $M_p$, the full dynamics approximation used for optimization ($A_{lin}$) will align closely with the actual dynamics. The figure demonstrates that the penalty for choosing an $M_p$ value too small, or neglecting it entirely, is greater than for picking a value that is larger than 580 kg. This is because, if $M_p$ were equal to zero, the actuator model's spring dynamics would alias when discretized. If $M_p$ approached infinity, this would equate to a model with infinite output impedance, which introduces error, but is a common modeling assumption for SEAs. Since the time step, $\Delta T$, and associated sampling frequency used to discretize the system in Section \ref{actuatordynamics} must be significantly greater than the largest eigenvector of the continuous system to avoid aliasing, the pseudo-mass modification is essential to allowing large time-steps, small linear program sizes, and fast run-times.

\section{Experiments}
\begin{figure}[tp]
\rule{0pt}{.25in}
\def\svgwidth{.98\columnwidth}
  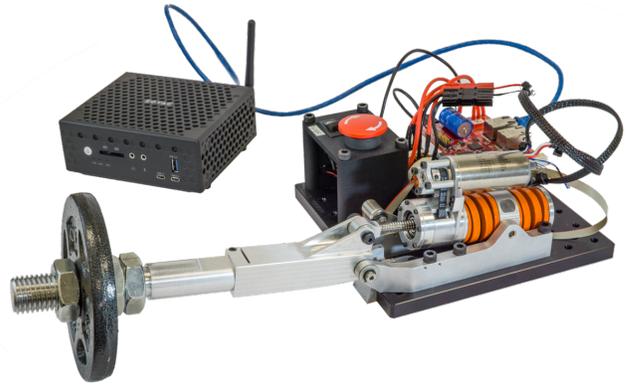
  \caption{The Taurus testbed, with an SEA whose spring is softer than Draco's viscoelastic elements by an order of magnitude.} 
  \vspace{-0.5cm}  \label{fig:taurus_and_eigenvalues}
\end{figure}

The findings from simulation were applied for validation on the single degree-of-freedom Apptronik Taurus testbed with the P170 Orion SEA (Fig. \ref{fig:taurus_and_eigenvalues}). The state space model in (\ref{eq:discrete}) was used with $p = 1$. 
Our trajectory optimization scheme relies on a well-identified model, so that the control system can depend heavily on the feed-forward, open loop command for high-speed tasks. System identification was performed using a least squares approach by fitting the parameters in (\ref{eq:nonlin1}) to the system's response to white noise and chirp signal current inputs. The parameter values are outlined in Table V. 
  \vspace{-0.4cm}  



\begin{table}[ht]
\rule{0pt}{.1in}

    \begin{minipage}{.5\columnwidth}
    \centering
    \label{P170ParametersTable}
      \caption{P170 Identified Parameters}
        \begin{tabular}{r|l}
\toprule
$M_S$ (kg) & 1\\
$k_S$ (N/m) & 698600\\
$\beta_S$ (Ns/m) & 500\\
$M_m$ (kg) & 250\\
$\beta_m$ (Ns/m) & 5885\\
$M_L$ (kg) & 0.227 \\
$M_p$ (kg) & 220 \\
$\beta_L$ (Ns/m) & 0\\
        \end{tabular}
    \end{minipage}%
    \begin{minipage}{.5\columnwidth}
    \centering
    \label{P170Constraints}
    \caption{P170 Constraints}
        \begin{tabular}{r|l}
\toprule
$\overline{\delta} $ (m) & 0.01\\
$z_{min}$ (m)& .0911  \\
$z_{max}$ (m) & .1389 \\
$\overline{\dot{y}}$ (m/s) & 0.3\\
$\overline{u}$ (A) & 3 \\
$\overline{\Delta {z}}$ (m) & 0.1 \\
$q_{N}$ (rad) & 1.57 \\
$z_N$ (m) & 0.11597 \\
$N$ & 105 \\
$\Delta T$ (s) & .005 \\
        \end{tabular}
    \end{minipage}%
      \vspace{-0.15cm}  
\end{table} 


\begin{figure}[tp]
\rule{0pt}{1.68in}
\def\svgwidth{.98\columnwidth}
  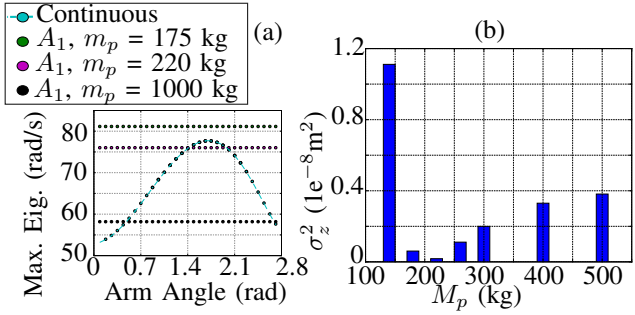
  \caption{Tuning pseudo-mass. (a) Maximum eigenvalue frequency for true arm-actuator system (Continuous) and various approximations ($A_1$) for the actuator alone with varying pseudo-mass ($M_p$). The arm points down at $0$ rad. Changing eigenvalue frequency for the true system is due to angle-dependent reflected inertia. The choice of $220$ kg is relatively accurate in the operational region centered around 1.57 rad. (b) Simulation error of discrete-time models used for trajectory optimization as a function of pseudo-mass, for a feedforward trajectory in the operational region, with a fixed time step. Average squared error relative to the trajectory of the true model. The low pseudo-mass example has aliasing errors. High pseudo-mass errors exist, but are not as extreme.}
  \label{fig:taurus_eigenvalues}
    \vspace{-0.5cm}
\end{figure}

\begin{figure*}[tp]
  \rule{0pt}{.2in}

\def\svgwidth{1.0\linewidth}
  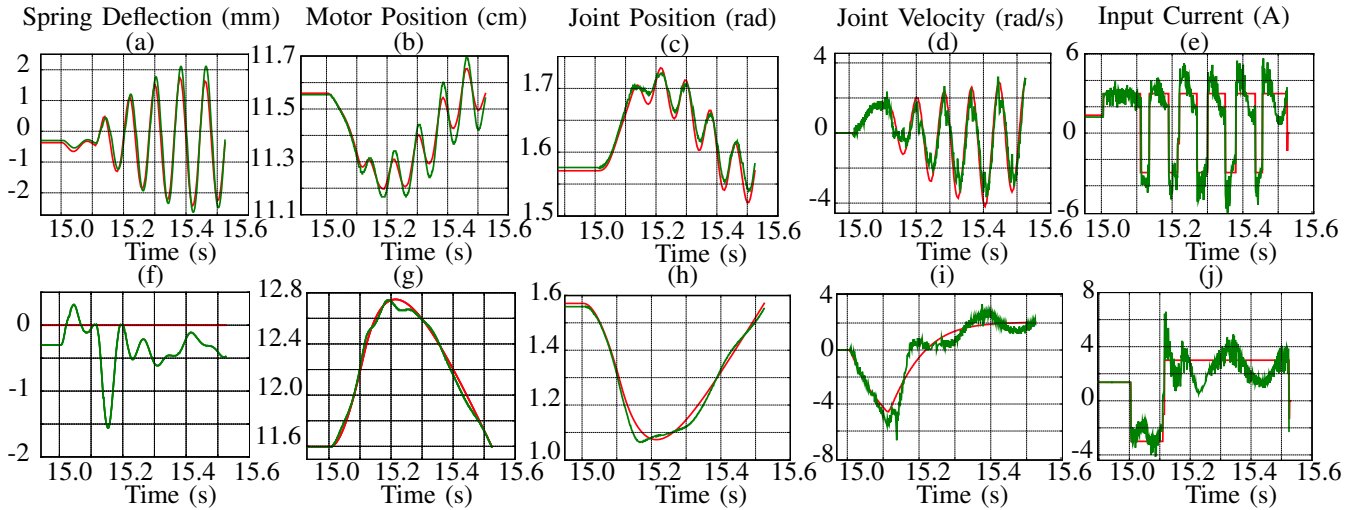
  \vspace{-0.6cm}
  \caption{(a)-(e) show the simulated (red) and actual (green) optimal behavior of the P170 actuator when spring dynamics are considered. The high-quality tracking in (a)-(e) support that the system has been well identified.  (f)-(j) show the expected and actual behaviors of the actuator when spring dynamics are not considered. These results demonstrate dynamic inconsistencies in tracking when the spring subsystem is neglected in planning. The experimental data are shifted by 5 ms to account for time delay. Fifteen seconds are used to interpolate to the starting position, and the optimal motion begins at 15 s.}
  \label{fig:experiment}
    \vspace{-0.25 cm}
\end{figure*}

The goal of the experiments was to maximize the actuator velocity in 0.52 seconds, as described by:

\begin{equation}\label{costfunction_p170}
\begin{aligned}
h(\mathbf{X}, \mathbf{U}) = &-\dot z_N
 +\sigma \sum_{n\in\mathcal N/N} u_{abs,i,n}, 
\end{aligned}
\end{equation}

\noindent where $\sigma$ equals $1e^{-8}$. 
In addition to our feedforward current command, we implemented a simple P controller, feeding back motor position. For increased stability we controlled motor position, rather than joint position, in order to control a collocated system from the control input \cite{kwak2017comparison}.
Because our configuration involves feedback, it is important for safety to ensure that conservative current limits are used in the optimization scheme, and that the software used for implementation upholds the hardware's actual upper limits. In this experiment an optimal trajectory was produced with a maximum allowable current of 3 A, but the motor saturation limit was 8 A. Constraints for optimization are shown in Table VI. A smaller time step was used in this optimization scheme to support the convergence of the specific problem. 

For one experiment, the optimal trajectory is devised with spring dynamics considered, and for comparison, an optimal trajectory is produced while ignoring spring states. Trajectory generation is first performed in simulation using CVXPY \cite{cvxpy} (to explore available solvers), to obtain the feedforward current command and desired trajectory. The difference in computational costs for the compliant and rigid systems are negligible in this case: four iterations and six seconds with compliance considered versus five iterations and six seconds without compliance. 



To select a reasonable pseudo-mass\footnote{Approximating reflected inertia of the arm w.r.t. actuator displacement.}, we plotted the maximum eigenvalues of the continuous system, and the maximum eigenvalue of $A_1$ for several distinct $M_p$ values. We chose $M_p=220$ kg based on the alignment of the largest eigenvalue over a range of likely arm configurations, as seen in Fig. \ref{fig:taurus_eigenvalues}.a. To quantify the error between the continuous dynamics and the approximated dynamics with a particular pseudo-mass, we obtained the $\mathbf{Z}$ trajectories for the continuous, true dynamics, $\mathbf{Z}_c$, and the approximate, linearized dynamics, $\mathbf{Z}_{lin}$, for a pre-defined input current trajectory in the expected operating region. The error associated with our pseudo-mass selection can then be expressed by the mean squared error of the $\mathbf{Z}_{lin}$ trajectory, $\sigma_z^2$, as seen in Fig. \ref{fig:taurus_eigenvalues}.b.

Fig. \ref{fig:experiment} shows the results of the experiments. To compare the expected and actual behaviors, the position states of the simulated and experimental data sets are filtered using a second order Butterworth filter with a cutoff frequency of 30 Hz. Table \ref{table:experiment_compare} summarizes the actual and ideal results. In both experiments, the actuator is able to start and end at the desired actuator position, .11597 m, with negligible error. With compliance modeled, the optimal trajectory is oscillatory in order to store and release energy, while the optimal trajectory for the rigid counterpart is a down-up motion. While there is $0.06\%$ error in the final velocity when compliance is considered, there is $8.21\%$ error in the final velocity when the system is considered rigid. Fig. \ref{fig:experiment} (e) and (j) show that the feedforward current aligns well with the actual required current when spring dynamics are considered, while there are more deviations from nominal when the system is considered rigid. When compliance is modeled, the ideal, final optimal velocity is $51\%$ greater than that achieved with the rigid model. These results demonstrate the benefit of modeling compliance for dynamically feasible motions, and the gains of leveraging compliance for high-performance tasks.   

\begin{table}[ht]
\caption{Experiment Results } 
\centering 
\begin{tabular}{c c c c} 
\hline\hline 
Configuration & Actual $z_N$ (m) & Ideal $\dot{z}_N$ (m/s) & Actual $\dot{z}_N (m/s)$ \\ [0.5ex] 
\hline 
Compliant & 0.11570 & 0.07671 & 0.07666 \\ 
Rigid & 0.11644 &  0.05094 & 0.04676 \\ 
\hline 
\end{tabular}
\label{table:experiment_compare} 
      \vspace{-0.15cm} 
\end{table}

\section{Discussion}

Actuator dynamics are often neglected from robot motion planning due to computational complexity, and our proposed method for trajectory optimization offers several advantages in this regard. First, directly capturing all relevant state and input constraints is an essential feature for a dynamically consistent trajectory. Our new robot--actuator interface, modified by pseudo-mass $M_p$, allows us to exploit the structural difference between a linear actuator admittance and a nonlinear robot impedance---which is novel and efficient. Through our formulation, we can increase the states of the system to include actuator dynamics without paying the computational cost typically associated with adding states to nonlinear optimization problems. 
Finally, we have demonstrated the gains in executing a high-performance task by leveraging compliance in the linear optimization subproblem. As actuators cannot function as perfect torque sources, planners that have the knowledge of the actuators' more-detailed abilities will allow them to produce achievable trajectories which can leverage the natural dynamics endowed by their low level components. 
    \vspace{-0.33cm}  







\addtolength{\textheight}{-12cm}   







\bibliographystyle{bib/IEEEtran}
\bibliography{bib/bib}

\begin{thebibliography}{10}
\providecommand{\url}[1]{#1}
\csname url@samestyle\endcsname
\providecommand{\newblock}{\relax}
\providecommand{\bibinfo}[2]{#2}
\providecommand{\BIBentrySTDinterwordspacing}{\spaceskip=0pt\relax}
\providecommand{\BIBentryALTinterwordstretchfactor}{4}
\providecommand{\BIBentryALTinterwordspacing}{\spaceskip=\fontdimen2\font plus
\BIBentryALTinterwordstretchfactor\fontdimen3\font minus
  \fontdimen4\font\relax}
\providecommand{\BIBforeignlanguage}[2]{{%
\expandafter\ifx\csname l@#1\endcsname\relax
\typeout{** WARNING: IEEEtran.bst: No hyphenation pattern has been}%
\typeout{** loaded for the language `#1'. Using the pattern for}%
\typeout{** the default language instead.}%
\else
\language=\csname l@#1\endcsname
\fi
#2}}
\providecommand{\BIBdecl}{\relax}
\BIBdecl

\bibitem{pratt1995series}
G.~A. Pratt and M.~M. Williamson, ``Series elastic actuators,'' in
  \emph{Intelligent Robots and Systems 95. `Human Robot Interaction and
  Cooperative Robots', Proceedings. 1995 IEEE/RSJ International Conference on},
  vol.~1.\hskip 1em plus 0.5em minus 0.4em\relax IEEE, 1995, pp. 399--406.

\bibitem{paine2014design}
N.~Paine, S.~Oh, and L.~Sentis, ``Design and control considerations for
  high-performance series elastic actuators,'' \emph{IEEE/ASME Transactions on
  Mechatronics}, vol.~19, no.~3, pp. 1080--1091, 2014.

\bibitem{sreenath2011compliant}
K.~Sreenath, H.-W. Park, I.~Poulakakis, and J.~W. Grizzle, ``A compliant hybrid
  zero dynamics controller for stable, efficient and fast bipedal walking on
  {MABEL},'' \emph{The International Journal of Robotics Research}, vol.~30,
  no.~9, pp. 1170--1193, 2011.

\bibitem{pratt2012capturability}
J.~Pratt, T.~Koolen, T.~De~Boer, J.~Rebula, S.~Cotton, J.~Carff, M.~Johnson,
  and P.~Neuhaus, ``Capturability-based analysis and control of legged
  locomotion, part 2: Application to m2v2, a lower-body humanoid,'' \emph{The
  International Journal of Robotics Research}, vol.~31, no.~10, pp. 1117--1133,
  2012.

\bibitem{vanderborght2013variable}
B.~Vanderborght, A.~Albu-Sch{\"a}ffer, A.~Bicchi, E.~Burdet, D.~G. Caldwell,
  R.~Carloni, M.~Catalano, O.~Eiberger, W.~Friedl, G.~Ganesh \emph{et~al.},
  ``Variable impedance actuators: A review,'' \emph{Robotics and autonomous
  systems}, vol.~61, no.~12, pp. 1601--1614, 2013.

\bibitem{chen2013optimal}
L.~Chen, M.~Garabini, M.~Laffranchi, N.~Kashiri, N.~G. Tsagarakis, A.~Bicchi,
  and D.~G. Caldwell, ``Optimal control for maximizing velocity of the
  compact™ compliant actuator,'' in \emph{Robotics and Automation (ICRA),
  2013 IEEE International Conference on}.\hskip 1em plus 0.5em minus
  0.4em\relax IEEE, 2013, pp. 516--522.

\bibitem{radulescu2012exploiting}
A.~Radulescu, M.~Howard, D.~J. Braun, and S.~Vijayakumar, ``Exploiting variable
  physical damping in rapid movement tasks,'' in \emph{Advanced Intelligent
  Mechatronics (AIM), 2012 IEEE/ASME International Conference on}.\hskip 1em
  plus 0.5em minus 0.4em\relax IEEE, 2012, pp. 141--148.

\bibitem{braun2012optimal}
D.~Braun, M.~Howard, and S.~Vijayakumar, ``Optimal variable stiffness control:
  formulation and application to explosive movement tasks,'' \emph{Autonomous
  Robots}, vol.~33, no.~3, pp. 237--253, 2012.

\bibitem{braun2013robots}
D.~J. Braun, F.~Petit, F.~Huber, S.~Haddadin, P.~Van Der~Smagt,
  A.~Albu-Sch{\"a}ffer, and S.~Vijayakumar, ``Robots driven by compliant
  actuators: Optimal control under actuation constraints,'' \emph{IEEE
  Transactions on Robotics}, vol.~29, no.~5, pp. 1085--1101, 2013.

\bibitem{werner2017optimal}
A.~Werner, B.~Henze, F.~C. Loeffl, S.~Leyendecker, and C.~Ott, ``Optimal and
  robust walking using intrinsic properties of a series-elastic robot,'' in
  \emph{IEEE-RAS International Conference on Humanoid Robots}, 2017.

\bibitem{werner2017generation}
A.~Werner, W.~Turlej, and C.~Ott, ``Generation of locomotion trajectories for
  series elastic and viscoelastic bipedal robots,'' in \emph{IEEE International
  Conference on Intelligent Robots and Systems}, 2017.

\bibitem{orekhov2015unlumped}
V.~L. Orekhov, C.~S. Knabe, M.~A. Hopkins, and D.~W. Hong, ``An unlumped model
  for linear series elastic actuators with ball screw drives,'' in
  \emph{Intelligent Robots and Systems (IROS), 2015 IEEE/RSJ International
  Conference on}.\hskip 1em plus 0.5em minus 0.4em\relax IEEE, 2015, pp.
  2224--2230.

\bibitem{schutz2016intuitive}
S.~Sch{\"u}tz, A.~Nejadfard, C.~K{\"o}tting, and K.~Berns, ``An intuitive and
  comprehensive two-load model for series elastic actuators,'' in
  \emph{Advanced Motion Control (AMC), 2016 IEEE 14th International Workshop
  on}.\hskip 1em plus 0.5em minus 0.4em\relax IEEE, 2016, pp. 573--580.

\bibitem{al05}
\BIBentryALTinterwordspacing
H.~Asada and J.~Leonard, \emph{2.12 Introduction to Robotics}.\hskip 1em plus
  0.5em minus 0.4em\relax Fall Massachusetts Institute of Technology: MIT
  OpenCourseWare, License: Creative Commons BY-NC-SA, 2005. [Online].
  Available: \url{https://ocw.mit.edu}
\BIBentrySTDinterwordspacing

\bibitem{koolen2016design}
T.~Koolen, S.~Bertrand, G.~Thomas, T.~De~Boer, T.~Wu, J.~Smith, J.~Englsberger,
  and J.~Pratt, ``Design of a momentum-based control framework and application
  to the humanoid robot atlas,'' \emph{International Journal of Humanoid
  Robotics}, vol.~13, no.~01, p. 1650007, 2016.

\bibitem{thomas2016towards}
G.~C. Thomas and L.~Sentis, ``Towards computationally efficient planning of
  dynamic multi-contact locomotion,'' in \emph{Intelligent Robots and Systems
  (IROS), 2016 IEEE/RSJ International Conference on}.\hskip 1em plus 0.5em
  minus 0.4em\relax IEEE, 2016, pp. 3879--3886.

\bibitem{cvx}
M.~Grant and S.~Boyd, ``{CVX}: Matlab software for disciplined convex
  programming, version 2.1,'' Mar. 2014.

\bibitem{kwak2017comparison}
S.~H. Kwak and S.~Oh, ``Comparison of resonance ratio control and inner force
  control for series elastic actuator,'' in \emph{Industrial Electronics
  Society, IECON 2017-43rd Annual Conference of the IEEE}.\hskip 1em plus 0.5em
  minus 0.4em\relax IEEE, 2017, pp. 7583--7588.

\bibitem{cvxpy}
S.~Diamond and S.~Boyd, ``{CVXPY}: A {P}ython-embedded modeling language for
  convex optimization,'' \emph{Journal of Machine Learning Research}, vol.~17,
  no.~83, pp. 1--5, 2016.

\end{thebibliography}

\end{document}